%% file: main.tex
\documentclass[10pt,twocolumn,letterpaper]{article}

\usepackage{iccv}
\usepackage{times,paralist}
\usepackage{epsfig}
\usepackage{graphicx}
\usepackage{amsmath}
\usepackage{amssymb,comment}

\usepackage[breaklinks=true,bookmarks=false]{hyperref}

\iccvfinalcopy 

\def\iccvPaperID{8010} 

\ificcvfinal\fi

\usepackage{xcolor}
\definecolor{bittersweet}{rgb}{1.0, 0.44, 0.37}
\definecolor{mygreen}{rgb}{0.29, 0.7, 0.48}

\newcommand{\gen}{\mathcal{L}}

\newtheorem{remark}{Remark}
\usepackage{caption} 
\usepackage[ruled, linesnumbered]{algorithm2e}
\usepackage[inline]{enumitem}

\begin{document}

\title{Neural TMDlayer: Modeling Instantaneous flow of features via SDE Generators}

\author{Zihang Meng\textsuperscript{\rm 1} ~~
                Vikas Singh\textsuperscript{\rm 1} ~~
                Sathya N. Ravi\textsuperscript{\rm 2} \\
\textsuperscript{\rm 1}University of Wisconsin-Madison\\
\textsuperscript{\rm 2}University of Illinois at Chicago\\
{\tt\small zihangm@cs.wisc.edu, vsingh@biostat.wisc.edu, sathya@uic.edu}
}

\maketitle
\ificcvfinal\thispagestyle{empty}\fi

\begin{abstract}
   We study how stochastic differential equation (SDE) 
   based ideas can inspire new modifications to existing 
   algorithms for a set of problems in computer vision. 
   Loosely speaking, our formulation is related to both 
   explicit and implicit strategies for data augmentation and group 
   equivariance, but is derived from new results in the SDE literature on estimating 
   infinitesimal generators of a class of stochastic 
   processes. If and when there is nominal agreement between 
   the needs of an application/task and the inherent 
   properties and behavior of the types of processes that we can efficiently handle, we obtain a very simple and efficient plug-in layer that can be incorporated 
   within any existing network architecture, with minimal modification and only a few additional 
   parameters. We show promising 
   experiments on a number of vision tasks 
   including few shot learning, point cloud transformers and deep 
   variational segmentation obtaining 
   efficiency or performance improvements.

\end{abstract}

\input{intro.tex}

\input{related.tex}

\input{method.tex}

\input{exp.tex}

\input{conclusion.tex}

{\small
\bibliographystyle{ieee_fullname}
\bibliography{egbib}
}

\end{document}

%% file: intro.tex
\section{Introduction}



Consider a deep neural network model with parameters $W$ which we train using the following update rule,
\begin{align}
	W \leftarrow W  -\eta \nabla_W\mathbb{E}_zR\left(W,z\right) \label{eq:sgd}
\end{align}
where $z$ is a random variable representing data and $R(\cdot)$ represents the loss function. Now, consider a slightly 
general form of the same update formula, 
\begin{align}
	W \leftarrow W  -\eta \nabla_W\mathbb{E}_zR\left(W,{\color{blue}T}z\right). \label{eq:gd-aug}
\end{align}
The only change here is the introduction of {\color{blue}$T$} which can be assumed to be {\em some} data transformation matrix. If $T=I$, we see that Stochastic Gradient Descent (SGD) is a special case of \eqref{eq:gd-aug} under 
the assumption that we 
approximate the expectation in \eqref{eq:gd-aug} with finite iid samples (or a mini-batch). 

Let us unpack the data transformation notation a bit to check what it offers.
If a set of transformations $T$ are chosen beforehand, and  
applied to the data samples before training commences, 
$Tz$ simply represents data samples derived via 
data augmentation. On the other hand, $Tz$ may not necessarily be 
explicitly instantiated as above.
For example, spherical CNN \cite{esteves17_learn_so_equiv_repres_with_spher_cnns} shows that 
when point cloud type data are embedded on the 
sphere with spherical convolutional operators, 
then it is possible to learn representations of data 
that are equivariant to the group action of rotations with {\bf no} explicit 
data augmentation procedure.
In particular, these approaches {\em register} each data point on a standard {\em template} (like the sphere) on which efficient convolutions can be defined based on differential geometric constructions -- in other words, utilizing the properties 
of the transformations $T$ of interest {\em and} how they relate the data points, such a treatment enables the updates 
to implicitly take into account the loss on $Tz$. 
Conceptually, many results \cite{esteves17_learn_so_equiv_repres_with_spher_cnns,spezialetti2019learning,qi2020learning} on equivariance 
show that by considering the {\em entire} orbit of 
each sample (a 3D point cloud) during training, for special types of $T$, 
it is possible to avoid explicit data augmentation.

We can take a more expanded 
view of the above idea. 
Repeated application of a transformation $T$ on data point 
$z$ produces a  discrete sequence $\{z(t)\}_{t=0}^{\infty}$ where $z(0)=z,z(t)=T^{t-1}z$. In general, the transformation matrix at 
the $t$-th step, denoted by $T(t)$, need not even 
be generated from a fixed matrix. Indeed, in practice $T(t)$ is selected from a set of appropriate transformations such as rotation, blur and so on, with some ordering, which 
could even be stochastic. 
At a high level, approaches such as 
\cite{esteves17_learn_so_equiv_repres_with_spher_cnns,cohen2018spherical} can be seen as a special case of \eqref{eq:gd-aug}. Making this argument 
precise needs adding an appropriate number of auxiliary variables and by averaging over all possible realizable $T$'s -- the specific steps are not particularly relevant since 
apart from helping set up the intuition we just described, algorithms for equivariance to specific group actions do not directly inform our development. 
For the sake of convenience, we will primarily focus on the continuous time system since under the same initial conditions,  the trajectories of both (continuous and discrete) systems coincide at all integers $t$.  

 \begin{figure}[!ht]
\centering
\includegraphics[width=0.475\textwidth]{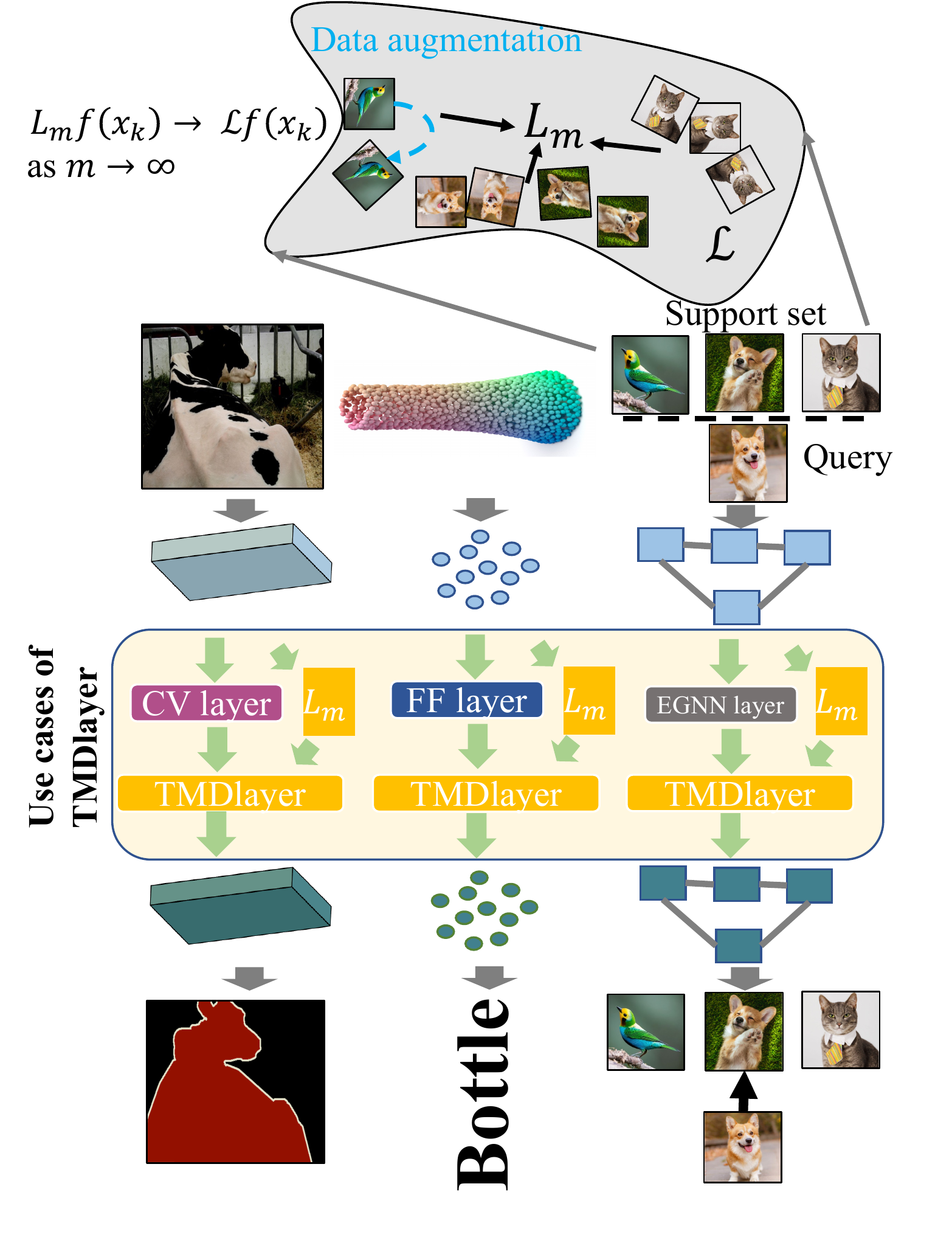} 
\vspace{-15pt}
\caption{\label{fig_overall_3tasks} \footnotesize Overview of TMDlayer use in few-shot recognition, point cloud learning and segmentation. ``EGNN'' refers to edge-labeling graph neural  network \cite{kim2019edge}; ``FF'' refers to feed-forward layer \eqref{eq:pt_transformer} and ``CV''  refers to our proposed 
	deep Chan Vese model \eqref{eq:cv_updating}. The manifold (top) describes the meaning of $\mathcal{L}$ and $L_m$: $\mathcal{L}$ captures the structure of the manifold. $L_m$ is an approximation of $\mathcal{L}$ constructed from samples. }
 \end{figure}

 {\bf What does $z(t)$ actually represent?} There are two interpretations of $z(t)$:
\begin{inparaenum}[\bfseries (i)]
\item it formalizes on-the-fly or instantaneous (smooth) data augmentation which are often used to accelerate training by exploiting symmetries in the landscape of $R$, and
\item a data dependent $T$ can be designed for invariance-like requirements,
  useful for downstream applications. In fact, learning data dependent transformations has also
  been explored by \cite{cubuk2019autoaugment}. 
\end{inparaenum}
  The starting 
point of this work is to exploit 
the view that the data sample provided 
to us is merely a {\em snapshot} 
of an underlying process which we will discuss shortly. 
Nonetheless, the key hypothesis is that 
specifying this process to 
our deep neural network model will be beneficial and provide a fresh 
perspective on some strategies that 
are already in use in the literature. 

{\bf Main ideas.} The foregoing use of ``process'' to describe the data 
sample hints at the potential use of an ordinary 
differential equations (ODE).
While ODE type constructions {\em can} be used to 
characterize simple processes, it will be 
insufficient to model more complex processes that will better reflect practical considerations. The key challenge in directly instantiating the ``$z(t)$''
idea 
for SDEs is that it is clearly infeasible since there are infinite possible trajectories for the same initial conditions. Our main insight is that recent results in 
the SDE literature  show that (under some technical conditions), the dynamics $z(t)$ can be {\bf completely} characterized by  (functions of) the infinitesimal generator $\gen$ of the process $z(t)$ which can be efficiently estimated using finite data. 
We exploit this result via 
a simple modification to the estimation procedure -- one that can be directly used within 
{\em any} backpropagation based training scheme.
{Specifically, we exploit the result from \cite{banisch2020diffusion} where the authors
  call the generator Target Measure Diffusion map (TMDmap).}
This leads to our {\bf TMDlayer} that 
can be conveniently dropped into 
a network, and be used as a plug-and-play module with {just a few additional parameters}. 
When utilized within standard deep learning pipelines, our layer allows incorporating much richer domain information  if available, or as a regularizer or an augmentation scheme, or as a substitute to an existing layer. 
We find this is beneficial to the overall performance of the model.

{\bf Our contributions.}
{
Models such as a Neural ODE \cite{chen2018neural} and Neural SDE \cite{liu2019neural} usually parameterize the dynamical system as a stand-alone model, and show 
how gradients can be efficiently backpropagated through this module. 
We take a different line of approach:
we propose a stochastic process inspired layer 
which, in its most rudimentary form, can be thought of as an augmentation scheme that can work with existing layers in deep neural networks. 
But different from explicit data augmentation (rotation, flipping) that happens in the 
input image space, our layer can be utilized in the feature space and is fully adaptive to the input.
But it is more than {\em another} augmentation
scheme. 
Our layer allows modeling 
the time varying/stochastic property of the data/features, and controls them by a proper parameterization which is highly parameter efficient. We show 
that this stochasticity is not only mathematically interesting, 
but can be exploited in applications including point cloud transformers, object segmentation and few-shot recognition.}

{\subsection{Related Work.}
Early work in vision has made extensive use of differential
equations \cite{chan2001active,malladi1995shape,rudin1994total,caselles1997geodesic}, especially for segmentation. 
In machine learning, differential equations are useful for manifold learning \cite{belkin2003laplacian} and semi-supervised learning \cite{belkin2006manifold,melas2020geometry} among others. 
Recently, a number of strategies combine differential equations with deep neural networks (DNNs) for solving vision problems.
For example, \cite{chen2017deeplab} utilizes a conditional random field after the CNN encoder to refine the semantic segmentation results whose update rules can be viewed as a differential equation
and \cite{marcos2018learning,hatamizadeh2019end} uses a CNN to extract visual features before feeding them to an active contour model which iteratively refines the contour according to the differential equation. Separately, the
literature includes strategies for
solving differential equations with DNNs \cite{kharazmi2019variational,michoski2020solving,li2020fourier}. 
Over the last few years, a number of formulations 
including neural ODE \cite{chen2018neural}, neural SDE \cite{liu2019neural} and
augmented neural ODE \cite{dupont2019augmented} have been proposed, motivated by the need to solve differential equation modules within DNNs. 
Note that \cite{liu2019neural} proposes to stabilize the neural ODE network with stochastic noise, which leads to a neural SDE, a setting quite 
different from the one studied here. Finally, we note that SDEs as a tool  
have also been used for stochastic analysis of DNNs \cite{chaudhari2018stochastic}.}

%% file: related.tex

%% file: method.tex
\section{Preliminaries} 
{\bf Background.} 
Partial differential equation (PDE) is a functional equation in which the solution satisfies given relations between its various partial derivatives interpreted as multivariable functions.
Consider a commonly used PDE model for segmentation -- the heat equation,$\frac{\partial u}{\partial t} = \Delta u,$ where $u$ depends on both $X$ and $t$. By the celebrated Feynman-Kac formula, we know that the solution $u$ can be equivalently written as a conditional expectation with respect to a  continuous time stochastic process $X_t$. This means that the solution (segmentation) $u$ can be obtained by averaging a sequence of stochastic integration problems. For prediction, we need an algebraic concept called the ``generator'' of a {\em function} (like a neural network) since we are more interested in the pushforward mappings  $f\left(X_t\right)$.

Given a time invariant stochastic process $X_t$, the (infinitesimal) generator $\gen$ of a function $f$  is defined as,\begin{align}
	\mathcal{L} f(X) := \lim\limits_{t\to 0} \frac{\mathbb{E}\left[f(X_t)\right]-f(X_0)}{t}.\label{eq:inf_gen}
\end{align}  If the process $X_t$ is deterministic, the expectation operator $\mathbb{E}$ becomes identity, and so the {\bf generator} $\gen$ simply measures the {\em instantaneous} rate of change in $f$ with respect to $X$.   In addition, say that $X_t$ can also be expressed as a (It\^{o}) Stochastic Differential Equation (SDE), i.e., $X_t$ satisfies:  \begin{equation}
	dX_t = b(X_t)dt + \sigma(X_t)dW_t,
	\label{ito_diff_0}
\end{equation}
where $W_t$ is a (multidimensional) Brownian motion with covariance $C$, and $b,\sigma$ represent drift and diffusion functions. Then, it turns out that $\gen$ can be written in closed form (without limit) as,\begin{align}
	\gen f= b\cdot \nabla f + \sigma C \sigma^T\cdot \nabla^2 f,
	\label{diff_ope_0}
\end{align}
where $\gen$ acts  as a {\em linear} operator on functions $f$, see \cite{kunita1997stochastic}. We will shortly explain how to estimate and utilize $\gen$.


{\bf Setup.} Consider the setting where $X$ represents our input features (say, an image as a 3D array for the RGB channels) and $f$ is a network with $L$ layers. Let the data be in the form of points $\mathcal{D}^{(m)}:={x_1,x_2,...,x_m}\in \mathcal{R}^N$ with $N>0$, which lie on a compact $d$-dimensional differentiable submanifold $\mathcal{M}\subseteq \mathcal{R}^N$ which is assumed to be unknown. We assume that $f$ in our case is defined implicitly using samples $x_i\in\mathcal{M}$, and so it is impossible to obtain closed form expressions for the operators $\nabla,\nabla^2$ in \eqref{diff_ope_0}. In such cases, recall that, when $\sigma\equiv0$, Diffusion maps \cite{coifman2006diffusion} uncovers the geometric structure $\mathcal{M}$ by using $\mathcal{D}^{(m)}$  to construct an $m\times m$ matrix $L_m$ as an approximation of the linear operator $\mathcal{L}$.  

{\bf Interpreting SDE.}
Recall that when $\mathcal{L}$ is used on the {\em input space}, it can model stochastic transformations to the input image (rotation and clipping are special cases). When $\mathcal{L}$ is used on the {\em feature space} (e.g., in an intermediate layer of a DNN), it can then model stochastic transformations of the features where it is hard to hand design augmentation methods. Moreover, it enables us to parameterize and learn the underlying stochastic changes/SDE of the features.

{\bf Roadmap.} In the next section, we describe the estimation of differential operator  $\mathcal{L}$ within deep network training pipelines. Based on this estimate, we define TMDlayer as a approximation to $f(X_{\Delta t}):=f(X,\Delta t)$ for a small time interval $\Delta t$ using Taylor's theorem.  In \S\ref{sec_applications}, we discuss four different applications of TMDlayer, where the pushforward measure  $f(X,\Delta t)$ under the flow of features (interpreted as a vector field) $f(X,0)$ may be a reasonable choice.

\section{Approximating $\mathcal{L}$ in Feedforward Networks}
We now discuss a recently proposed nonparametric procedure to estimate $L_m$ given finite samples $x$ when $\sigma\not\equiv0$. This is an important ingredient because in our setup, we often do not have a meaningful model of minibatch samples, especially in the high dimensional setting (e.g., images).

{\bf Constructing $L_m$ in DNN training.}
The definition in \eqref{eq:inf_gen} while intuitive, is not immediately useful for computational purposes.  Under some technical conditions such as smoothness of $b,\sigma,f$, and the rank of $C$, \cite{banisch2020diffusion} recently showed that for processes that satisfy \eqref{ito_diff_0},  it is indeed possible to construct finite sample estimators $L_m$ of $\gen$. {In \cite{banisch2020diffusion}, the approach is called Target Measure Diffusion (TMD) so we call our proposed layer, a TMDlayer. }

To construct the differential operator, we first need to  compute a kernel matrix $K\in R^{m\times m}$ from the data. For problems involving a graph or a set of points as input, we can simply use the given data points ($m$ would be the number  of nodes in the graph, or the number of points in the set),  while for problems with a single input (e.g., standard image classification), we may not have access to $m$  data points directly. In this case, we can construct the kernel matrix by sampling a batch from the dataset and  processing them together because  we can often assume that the entire dataset is, in fact, sampled from some underlying distribution. 

\begin{algorithm}[!ht]
  \SetAlgoLined
  {\footnotesize
 \textbf{Input:} Function $f$, a batch of data samples $X=\{x_1,...,x_m\}$,  coefficient $\epsilon$, parameterized time interval $\Delta t$  \\
 Construct distance matrix $K_\epsilon$ by \eqref{eq:kernel} \\
 Compute kernel density estimate: $q_\epsilon(x_i)=\sum_{j=1}^m(K_{\epsilon})_{ij}$\\
 Parameterize target distribution by \eqref{eq:pi}\\
 Form the diagonal matrix $D_{\epsilon,\pi}$ with components $(D_{\epsilon,\pi})_{ii}=\pi^{1/2}(x_i)q^{-1}_{\epsilon}(x_i)$\\
 Use $D_{\epsilon,\pi}$ to right-normalize $K_\epsilon$: $K_{\epsilon,\pi} = K_\epsilon D_{\epsilon_\pi}$\\
 Construct $L_m$ by \eqref{eq:Lm} using $(\tilde{D}_{\epsilon,\pi})_{ii} :=\sum_{j=1}^m(K_{\epsilon,\pi})_{ij}$\\
 \textbf{Return:} $f(X) + \Delta t\cdot L_mf(X)$
  }
 \caption{Operational steps in TMDlayer}
 \label{alg}
\end{algorithm}

After getting the set of data samples, we first project the  data into a latent space $\mathbb{R}^h$ with a suitable $h$ using a learnable linear layer, before evaluating them with an appropriate kernel function such as,
\begin{equation}
\label{eq:kernel}
k_\epsilon(x,y) = \exp(-(4\epsilon)^{-1}\|x-y\|^2).
\end{equation}
We then follow \cite{banisch2020diffusion} to construct the differential operator $\mathcal{L}$ as follows: we compute the kernel density estimate $q_\epsilon(x_i)=\sum_{j=1}^m(K_\epsilon)_{ij}$. Then, we form the diagonal matrix $D_{\epsilon, \pi}$ with  components $(D_{\epsilon,\pi})_{ii}=\pi^{(1/2)}(x_i)q_\epsilon^{-1}(x_i)$. Here, we allow the network to learn $\pi$ by 
\begin{equation}
\label{eq:pi}
	\pi^{1/2}(x_i) = g(x_i),
\end{equation}
where $g$ can be a linear layer or a MLP depending on the specific application. Next we use $D_{\epsilon, \pi}$ to right-normalize the kernel matrix $K_{\epsilon,\pi}=K_\epsilon D_{\epsilon, \pi}$ and use $\tilde{D}_{\epsilon, \pi}$ which is the diagonal matrix of row sums of $K_{\epsilon, \pi}$ to left-normalize $K_{\epsilon, \pi}$.
Then we can build the TMDmap operator as
\begin{equation}
	L_{m} = \epsilon^{-1}(\tilde{D}_{\epsilon,\pi}^{-1}K_{\epsilon, \pi}-I).\label{eq:Lm}
\end{equation}
We will use \eqref{eq:Lm} to form our TMDlayer as described next.

\subsection{TMDlayer: A Transductive Correction via $L_m$}
Observe that \eqref{ito_diff_0} is very general and can represent many computer vision tasks where the density $\pi$ could be defined using a problem specific energy function, and $W_t$ is the source of noise. In other words, we aim to capture the underlying structure of the so-called image manifold \cite{zhu2016generative} by using its corresponding differential operator \eqref{diff_ope_0}. Intuitively, this means that if we are provided a network $f_W$ with parameters $W$, then by Taylor's theorem, the infinitesimal generator estimate $L_m$ can be used to approximate the change of $f_W$ as follows:
\begin{equation}
	\mathbb{E}_xf_W(x, \Delta t) \approx f_W(x,0) + \Delta t \cdot {L}_m [f_W], \label{eq:taylor}
\end{equation}
where $[f_W]\in\mathbb{R}^m$ such that the $i-$th coordinate $[f_W]_i=f_W(x_i)$, and $\Delta t$ is interpreted as a hyperparameter in our use cases, see Algorithm \ref{alg}.

{\bf Inference using $L_m$.}
In the ERM framework, typically, each test sample is used independently, and identically i.e.,  network (at optimal parameters) is used in a sequential manner for predictive purposes. Our framework allows us to further use relationships between the test samples for prediction. In particular, we can design custom choices of $b,\sigma$ tailored for downstream applications. For example, in applications that require robustness to {\em small} and {\em structured} perturbations, it may be natural to consider low bias diffusion processes i.e., we can prescribe the magnitude using $\|b\|_p\leq \kappa$ almost everywhere for some small constant $\kappa>0$ (akin to radius of perturbation)   and structure using diffusion functions $\sigma, C$. Inference can then be performed using  generators $\gen$ derived using the corresponding process.

{\bf Layerwise $\gen$ for improved estimation of $L_m$.} While \eqref{eq:taylor} allows us to use $L_m$ for {\em any} network with no modifications, using it naively can be unsatisfactory in practice. For example, often we find that  features from input layers might not be too informative for the task and may hinder training, especially in the early stages. We suggest a simple adjustment: instead of applying approximation in \eqref{eq:taylor} on the entire network, we do it layerwise -- this could be every intermediate layer or several layers of interest. It means that $f$ can in principle be any layer (e.g., a layer in graph neural networks or a layer in Resnet), as shown in Fig. \ref{fig_overall_3tasks}. 

{\bf Justification.} Recall that most feed-forward neural networks can be completely defined by a finite sequence of linear transformations followed by activation functions (along with intermediate normalization layers). One option is to estimate $L_m$ by directly applying the Taylor series-like expansion in \eqref{eq:taylor} on $f=f^l\circ f^{l-1}\circ\cdots f^1$ where $l$ represents the number of layers.  However, from \eqref{eq:taylor} we can see that the variance of such an estimate of the value $L_m[f_W]$  will be high due to the well-known propagation of uncertainty phenomenon (across $f^i$'s). To avoid this, we can estimate $L_m[f_W]$ in a sequential manner i.e., use  $L_m[f^{i-1}_W]$ to estimate $L_m[f^{i}_W]~\forall~ i\in[l]$. We will show in \S\ref{sec_applications} that this parameterization can be useful in various applications.

{{\bf Synopsis.}
We briefly summarize the benefits of our TMDlayer next: \begin{enumerate*}[label=(\roman*)]
\item Our TMDlayer can parameterize the underlying stochastic transformations of features, providing a way to augment features at any layer. 
\item The stochasticity/randomness in our TMDlayer is a stability inducing operation for robust predictive purposes \cite{hardt2016train}. 
\item Our TMDlayer is parameter efficient. All we need is a projection
linear layer $h$ and a linear layer $g$ parameterizing the density $\pi$ and a
 scalar parameter $\Delta t$.  In practice, {we can
 work with a small latent dimension (e.g., $h$ = 16) when constructing $L_m$, thus the
 total number of parameters in  TMDlayer is very small}
 when compared with the layer function $f$ in most deep learning
 applications.
 \end{enumerate*}
But the reader will see that
a mild limitation of the SDE perspective in practice is that, in principle, 
the dynamics may eventually get stuck in a meta-stable state. This means that in this case, the estimate $L_m$ will not be very informative in the forward pass, and so the gradient estimates might be biased. In such cases, it may be useful to add points by sampling on the orbit if needed. We will now describe four different vision settings where our TMDlayer can be instantiated in a plug-and-play manner.

%% file: exp.tex
\section{Applications}
\label{sec_applications}
In this section, we evaluate our TMDlayer in the context of different applications. As a warm-up, in \S\ref{resnet-exp}, 
we demonstrate the use of TMDlayer on 
a simple image classification task. We study its properties in both inductive and transductive settings. 
Then, in \S\ref{sec:point_cloud}, we move to learning with point cloud datasets. Here, we see that the data type naturally 
offers a suitable object for leveraging the features of TMDlayer. In this case, we conduct experiments in an inductive setting. 
Next, in \S\ref{sec:object_segmentation}, we explore the use of TMDlayer on a segmentation task (also in an inductive setting). 
We propose a novel deep active contour model which can be viewed as a dynamical process within a neural network. We 
demonstrate the use of our TMDlayer on top of such a dynamical process. Finally, in 
\S\ref{sec:few_shot}, 
we investigate few-shot learning. Here, the problem setup 
natively provides the graph needed for computing our $L_m$ and allows transductive inference.

\subsection{A Simple Sanity check on Resnet}
\label{resnet-exp}
We start with a simple example of image classification on CIFAR10 \cite{krizhevsky2009learning}
using Resnet \cite{he2016deep}, to demonstrate applicability of our
TMDlayer and evaluate its behavior.


\subsubsection{Role of TMDlayer: {Finetuning/Robustify} Resnet}
We choose Resnet-18 as the backbone network and simply treat each of its three residual blocks Res as $f$ (see \cite{he2016deep} for details of a residual block) in TMDlayer as follows,
\begin{equation}
	\label{eq:residual-block}\nonumber
    f(x^{l-1}) = \text{Res}(x^{l-1})\implies  x^{l} = f(x^{l-1}) + \Delta t\cdot L_m f(x^{l-1}),
\end{equation}
 where $x^l$ is the feature at the $l$-th layer and $L_m$ is constructed from a mini-batch of samples.
 
\subsubsection{Experimental results}
During training, we first
sample $m$ data points in a batch and use it as the input so that we can construct $L_m$. {During test time, an input batch also contains $m$ samples (similar to training time)}, where $m$ increases from $1$ to $200$. We can see from Table \ref{table:resnet} that $m$ {\em does} have an influence on the test accuracy where a larger $m$ performs better than a smaller $m$. A key reason is that $L_m$ using a larger $m$ can better capture the geometric structure of the data. 

We also test whether our TMDlayer can help improve the robustness of the network. We can assess this property by adding random noise to the 
input image and evaluating the test accuracy (see Table \ref{table:resnet_noise}). With our TMDlayer, the network is more noise resilient. 
This can be partly attributed to the use of our parameterized $\Delta t$, which allows the network to control the stochastic process in the TMDlayer adaptively and dependent on the input. In summary,
the performance profile is similar (Tab. \ref{table:resnet}) with small improvements in robustness (Tab. \ref{table:resnet_noise}). 

 \begin{table}[!h]\small
\centering
\begin{tabular}{l  | c | c }
\hline
$m$ &  Inference w/ TMDlayer  & Accuracy (\%)\\
\hline
  1 & No & 75.15\\
  1 & Yes & 87.35\\
  10 & Yes & 87.65 \\
  50 & Yes & 88.14 \\
  100  & Yes  &  88.52  \\
  150 & Yes & 88.55\\
  200 & Yes & 88.25\\
  \hline
\end{tabular}
\caption{\small Accuracy on test set of CIFAR10 after adding TMDlayer to Resnet-18. Here, $m$ is the batch size used to construct $L_m$ during test/inference time. The accuracy of Resnet-18 (trained/tested without TMDlayer) is $88.27\%$ comparable to $m \in \{50,100,150\}$.}
\label{table:resnet}
\end{table}

 \begin{table}[!h]\small
\centering
\begin{tabular}{l  | c | c | c | c | c }
\hline
$\sigma$  &  0.01  & 0.02 & 0.03 & 0.05 & 0.1  \\
\hline
  Resnet-18   & 87.54 & 83.90 & 75.85 & 53.87 & 17.27 \\
  Ours  & 87.79 & 84.37 & 77.96 & 56.18 & 19.18\\
  \hline
\end{tabular}
\caption{\small Accuracy on CIFAR10 when adding random noise (mean = $0$, std = $\sigma$) to the input. ``Ours'' refers to Resnet-18  plus the TMDlayer.}
\label{table:resnet_noise}
\vspace{-10pt}
\end{table}

 \subsection{Point cloud transformer}
 \label{sec:point_cloud}
 Tasks involving learning with point cloud data is important within 3D vision. The input here is usually a 3D point cloud represented by a set of points, each 
 associated with its own feature descriptor. These points can be naturally thought of as samples from the underlying distribution which captures the 
 geometric structure of the object. The problem provides an ideal 
 sandbox to study the effect of our TMDlayer. 
 But before we do so, we provide some context for where and 
 how the TMDlayer will be instantiated. 
 Recently, \cite{guo2020pct} proposed a transformer based model for point cloud learning which achieves state-of-the-art performance on this task -- and 
 corresponds to an effective and creative use of transformers in this setting. 
 Nonetheless, Transformer models are known to be parameter costly (e.g., see \cite{beltagy2020longformer,xiong2021nystr,zeng2021you} for cheaper approximations effective in NLP settings)
 and it is sensible to check to what extent our TMDlayer operating on a simple linear layer can be competitive with the transformer layer proposed in  
\cite{guo2020pct}. Our goal will be to check if significant parameter efficiency is possible.
 
 \subsubsection{Problem formulation}
 Denote an input point cloud $\mathcal{P}\in R^{N\times d}$ with $N$ points, each with a $d$-dimensional feature descriptor. The classification task is to predict a class or label for the entire point cloud.
 
 \subsubsection{Role of TMDlayer: Replace transformer layer}
 The point cloud transformer layer in \cite{guo2020pct} is constructed as,
 \begin{equation}
 	F_{out} = FF(F_{in}-F_{sa}) + F_{in},
 	\label{eq:pt_transformer}
 \end{equation}
 where FF refers to their feed-forward layer (a combination of Linear, BatchNorm and ReLU layer), and $F_{sa}$ is the output of self-attention module which takes $F_{in}$ as an input (we refer the reader to \cite{guo2020pct} for more details of their network design, also included in our appendix). 
 
 A Transformer layer is effective for point cloud because it simultaneously captures the relation between features of all points. Since our TMDlayer can be viewed as a diffusion operator which
 captures the structure of the underlying data manifold from the data, we can check to what extent its ability suffices.
 We use the TMDlayer on a single feed-forward layer to replace the Transformer layer in \eqref{eq:pt_transformer}. 
 \begin{equation}
 	F_{out} = FF(F_{in}) + \Delta t\cdot L_m FF(F_{in}).
 	\label{eq:pt_tmd}
 \end{equation}
 
 Surprisingly, it turns out that this simple layer can perform comparably with the carefully designed Transformer layer in \eqref{eq:pt_transformer} while offering a much more favorable parameter efficiency profile. Here, $L_m$ is constructed using points of the same point cloud (setting is identical to baselines).
 
 
 \subsubsection{Experimental results}
 
 \textbf{Dataset.} We follow \cite{guo2020pct} to conduct a point cloud classification experiment on ModelNet40 \cite{wu20153d}. The dataset contains $12311$ CAD models in $40$ object categories, 
 widely used in benchmarking point cloud shape classification methods. We use the official splits for training/evaluation. 
 
 \textbf{Network architecture and training details.} We use the same network as \cite{guo2020pct} except that we replace each point cloud transformer layer with a TMDlayer built on a single feed forward layer. We follow \cite{guo2020pct} to use the same sampling strategy to uniformly sample each object via $1024$ points and the same data augmentation strategy during training. The mini-batch size is $32$ and we train $250$ epochs using SGD (momentum $0.9$, initial learning rate $0.01$, cosine annealing schedule). The hidden dimension is $256$ for the whole network and
 $16$ for constructing $L_m$ (in TMDlayer).
  
 \textbf{Results.}
 We see from Table \ref{table:point_cloud_classfication} that our approach achieves comparable performance with \cite{guo2020pct}. In terms of the number of parameters,
using hidden dimension $256$  (used in this experiment) as an example, one self-attention layer contains $148$k parameters; one linear layer contains $65.5$k parameters; and the TMDlayer module only needs $4$k parameters.

  \begin{table}[!h]\small

\centering
\begin{tabular}{l  | c | c |c  }
\hline
Method & Input & \#Points & Accuracy(\%)\\
\hline
PointNet \cite{qi2017pointnet} & P & 1k & 89.2\\
A-SCN \cite{xie2018attentional} & P & 1k & 89.8 \\
 SO-Net \cite{li2018so} & P, N & 2k & 90.9\\
  Kd-Net \cite{klokov2017escape} & P & 32k & 91.8 \\
  PointNet++ \cite{qi2017pointnet}  & P & 1k & 90.7\\
  PointNet++ \cite{qi2017pointnet} & P, N & 5k & 91.9 \\
 PointGrid \cite{le2018pointgrid}  & P & 1k & 92.0 \\
 PCNN \cite{atzmon2018point}  & P & 1k & 92.3 \\
  PointConv \cite{wu2019pointconv} & P, N & 1k & 92.5\\
 A-CNN \cite{komarichev2019cnn} & P, N & 1k & 92.6\\
 DGCNN \cite{wang2019dynamic} & P & 1k & 92.9 \\
PCT \cite{guo2020pct} & P & 1k & \textbf{93.2}\\
\textbf{Ours} & P & 1k & 93.0\\
\hline
\end{tabular}
\caption{\small Classification results on ModelNet40. Accuracy means overall accuracy. P = points, N = normals. “Ours” means  replacing transformer layers in PCT with TMDlayer.}
\label{table:point_cloud_classfication}
\vspace{-10pt}
\end{table}

\subsection{Object segmentation}
\label{sec:object_segmentation}
Here, we show that our TMDlayer (a dynamical system) can also be built on top of another dynamical system. 
We do so by demonstrating experiments on object segmentation.

Recall that active contour models are a family of effective segmentation models which evolve the contour iteratively until a final result is obtained.
Among many options available in the literature (e.g., \cite{ronfard1994region,subakan2011quaternion,xu2000image}), the widely used Chan-Vese \cite{chan2001active} model evolves the contour based on a variational functional.  Here, we propose to combine the Chan-Vese functional with a deep network by parameterizing the iterative evolution steps and build our TMDlayer on top of it. We see that this simple idea leads to improved results. The appendix includes more details of our model. 

\subsubsection{Problem formulation}
Let $\Omega$ be a bounded open subset of $R^2$, where $\partial \Omega$ is its boundary. Let $I: \bar{\Omega}\rightarrow R$ be an image, object segmentation involves predicting
a dense map in $\bar{\Omega}\rightarrow 0/1$ where $1$ (and $0$) indicates the object (and background). In our formulation, we parameterize  the object contour
by a level set function $\phi:\Omega\to\mathbb{R}$ and evolve it within 
the DNN. {We note that hybrid approaches using level sets together with DNNs is not unique 
to our work, see \cite{marcos2018learning,yuan2020deep}}.



\subsubsection{Role of TMDlayer: in deep active contour model}



Our proposed deep active contour model evolves the contour in the form of a level set function within the network, and the update scheme is,
\begin{equation}
\phi^l = \phi^{l-1} + \frac{\partial \phi}{\partial t}\Delta t', 
\label{eq:cv_updating}
\end{equation}
where $\phi^{l-1}$ is the level set function at layer $l-1$ and $\frac{\partial \phi}{\partial t}$ is derived from our proposed deep variational functional. The appendix includes more details of our model, the variational functional, and the derivation of update equation.

Denote the update function in \eqref{eq:cv_updating} as $\phi^l=f(\phi^{l-1})$. Then, our TMDlayer forward pass can be written as, 
\begin{equation}
	\phi^{l} = f(\phi^{l-1}) + \Delta t\cdot L_m f(\phi^{l-1}).
	\label{eq:tmd_segmentation}
\end{equation}

\begin{remark}
Note that $\Delta t'$ in \eqref{eq:cv_updating} and the $\Delta t$ in \eqref{eq:tmd_segmentation} correspond to two different dynamical systems. The first one pertains to the update function of the deep active contour model and the second one
refers to the TMDlayer. $L_m$ in \eqref{eq:tmd_segmentation} is constructed using samples from the same mini-batch.
\end{remark}

\begin{remark}
	Note that our proposed segmentation model is different from \cite{yuan2020deep} which uses the variational energy function directly as the final loss, whereas we are parameterizing the updating steps within our network so that the final output will already satisfy low variational energy.  
\end{remark}

%
%

\subsubsection{Experimental results}

\textbf{Dataset.} 
The Vaihingen buildings dataset consists of 168 building images extracted from the training set of ISPRS ``2D semantic labeling contest'' with a resolution of 9cm. We use only 100 images to train the model and the remaining 68 serve as the test set.

\textbf{Network Architecture and Experiment Setup.}
\label{building_setup}
We use an encoder CNN with an architecture similar to \cite{hariharan2015hypercolumns} and \cite{marcos2018learning}. The input is the original image. The network is trained with a learning rate of $10^{-4}$ for 300 epochs using a batch size of $10$. We setup our baseline using the same CNN architecture to predict the segmentation mask without our Chan-Vese update module. Previous works combining active contour model and deep learning \cite{marcos2018learning,ling2019fast} often can only be used to provide segmentations of a single building based on manual initialization or another initialization (based on a separate algorithm) whereas our model can be used to segment multiple buildings in the image {\em without} any initialization. So, the results cannot be meaningfully compared. See our appendix for more details about the setup.



 \begin{figure}[!t]
\centering
\includegraphics[width=0.45\textwidth]{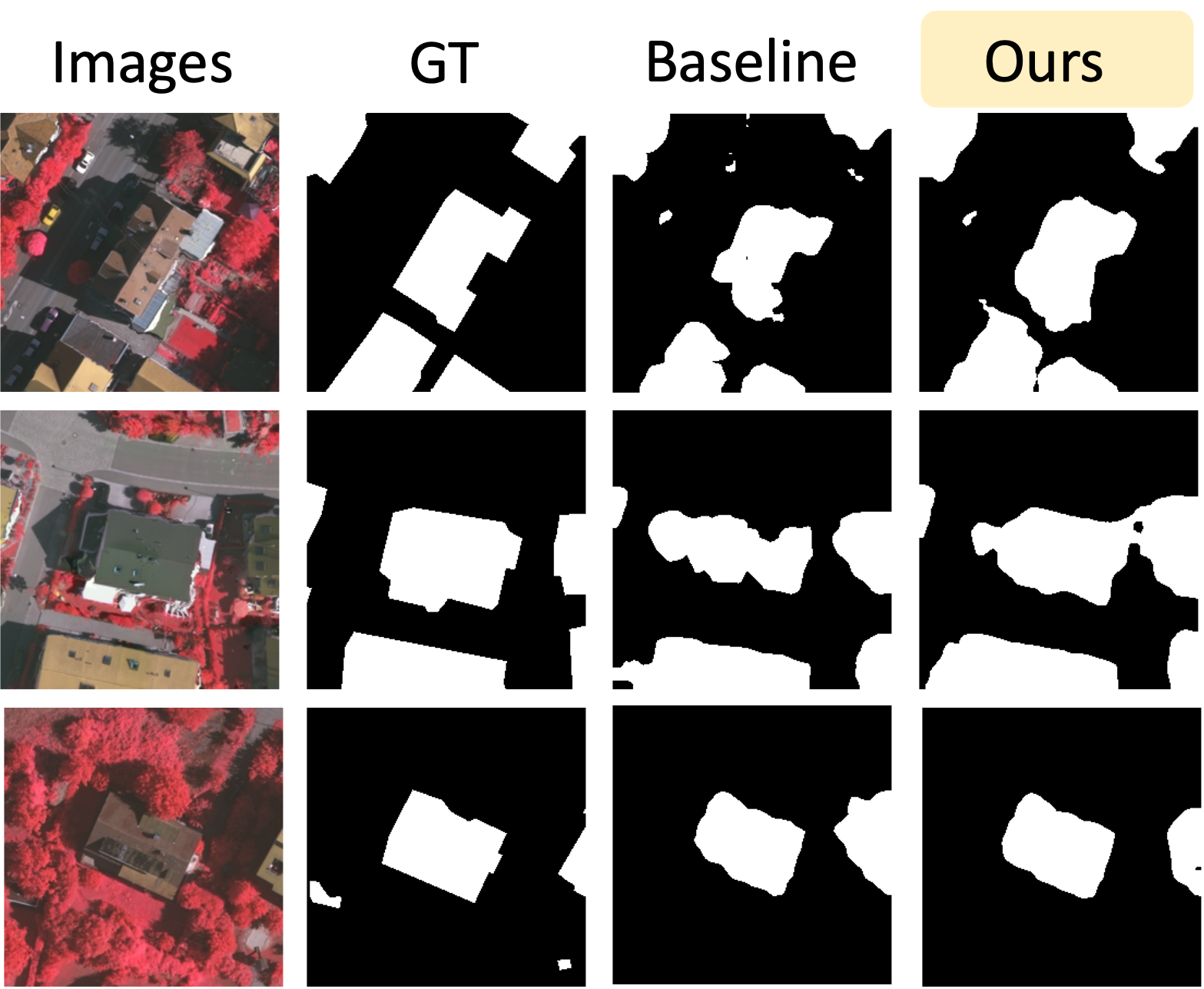} 
\vspace{-8pt}
\caption{\label{fig_qualitative_building} \small Qualitative results on Vaihingen dataset. Our model performs well despite the small sample size.}
\vspace{-6pt}
 \end{figure}



\textbf{Results and Discussion.}
We use the average Intersection over Union (IoU) to evaluate the performance on Vaihingen dataset: the baseline yields $\textbf{68.9}$ while our model without TMDlayer achieves $\textbf{73.5}$ and our complete model with TMDlayer achieves $\textbf{74.6}$, which is a significant improvement in terms of IoU.
This experiment shows that our TMDlayer can be built on top of another dynamical system and can provide additional benefits. 
%
%
%
Qualitative results of the baseline and our model are shown in Fig. \ref{fig_qualitative_building}. Our method tends to predict a more precise shape/boundary, and also fixes some flaws/errors relative to the baseline results.

 \subsection{Few-shot learning}
 \label{sec:few_shot}
 In $N$-way $B$-shot few-shot learning, the input is a set of
 $N \times B$
 samples which naturally forms a fully connected graph. This serves 
 to construct the differential operator $L_m$. To provide context for where and how 
 our TMDlayer will be instantiated, we note that 
 \cite{kim2019edge} proposed a GNN approach (EGNN) for few-shot learning and this model achieves state-of-the-art performance. We
 show that by adding our TMDlayer, the performance increases by a clear margin.
 
\subsubsection{Problem formulation}
  Few-shot learning classification seeks to learn a classifier given only a few training samples for every class. Each few-shot classification task $\mathcal{T}$ contains a support set  $S$ which is a  set of labeled input-label pairs and a query set $Q$ (an unlabeled set where the learned classifier is evaluated). Given $B$ labeled samples for each of $N$ classes in the support set $S$, the problem is a $N$-way $B$-shot classification problem.
 
 \subsubsection{Role of TMDlayer: Use in graph neural network}
  Let $\mathcal{G}$ be the graph formed by samples from the task $\mathcal{T}$, with nodes denoted as $\mathcal{V}:=\{V_i\}_{i=1,...,|\mathcal{T}|}$. The node feature update equation is designed as (we refer readers to \cite{kim2019edge} or our appendix for more details about the network)
 \begin{equation}
 	v_i^l = \text{NodeUpdate}(\{v_i^{l-1}\}, \{e_{ij}^{l-1}\}; \theta_v^l),
 	\label{EGNN_update}
 \end{equation}
 where $v_i^l$ is the feature of node $i$ at $l$-th layer, $e_{ij}$ is the edge feature between node $i$ and node $j$, and $\theta$ refers to the parameters in the update function. We abstract \eqref{EGNN_update} as $v_i^l=f(v_i^{l-1})$ and use our TMDlayer as,
 \begin{equation}
 	v_i^l = f(v_i^{l-1}) + \Delta t\cdot L_m f(v_i^{l-1}).
 	\label{eq:tmd_fewshot}
 \end{equation}
 
 \begin{remark}
 In \eqref{eq:tmd_fewshot}, the $L_m$ is constructed using samples from the same episode, and $f$ is a GNN module updating the node features using all node features and edge features.
 \end{remark}
 
 \subsubsection{Experimental results} 
 
 \textbf{Dataset.} We follow \cite{kim2019edge} to conduct experiments on {\it mini}ImageNet, proposed by \cite{vinyals2016matching} and derived from ILSVRC-12 dataset \cite{russakovsky2015imagenet}. Images are sampled from $100$ different classes with $600$ samples per class (size $84\times 84$ pixels). We use the same splits as in \cite{ravi2016optimization,kim2019edge}: $64$, $16$ and $20$ classes for training, validation and testing respectively.
 
 \textbf{Network architecture and training details.}
 We use the same graph neural network architecture and follow the training strategy as in \cite{kim2019edge} by utilizing the code provided by the authors. We add our TMDlayer as shown in \eqref{eq:tmd_fewshot} to each node update layer in the graph neural network, with a latent dimension of $16$ for constructing $L_m$. 
 We follow \cite{kim2019edge} to conduct experiments for 5-way 5-shot learning, in both transductive and non-transductive settings, as well as for both supervised and semi-supervised settings.
 The network is trained with Adam optimizer with an initial learning rate of $5\times 10^{-4}$ and weight decay of $10^{-6}$. The learning rate is cut in half every $15000$ episodes. For evaluation, each test episode was formed by randomly sampling $15$ queries for each of $5$ classes, and the performance is averaged over $600$ randomly generated episodes from the test set. Note that the feature embedding module is a convolutional neural network which consists of four blocks (following \cite{kim2019edge}) and used in most few-shot learning models without any skip connections. Thus,  Resnet-based models are excluded from the table for a fair comparison. We refer the reader to \cite{kim2019edge} or the appendix for more training and evaluation details.
 
 \textbf{Results.}
 The performance of supervised and semi-supervised 5-way 5-shot learning is given in Tables \ref{table:few_shot}--\ref{table:semi_few_shot} respectively.
 Our TMDlayer leads to consistent and clear improvements in both supervised and semi-supervised settings (also for transductive/non-transductive settings).

 \begin{table}[!ht]\small

\centering
\begin{tabular}{l  | c | c }
\hline
Model & Trans. & Accuracy(\%)\\
\hline
 Matching Networks \cite{vinyals2016matching}  & No  &  55.30  \\
 Reptile \cite{nichol2018first}  & No  &  62.74  \\
 Prototypical Net \cite{snell2017prototypical}  & No  &  65.77  \\
 GNN \cite{garcia2017few}  & No  & 66.41   \\
  EGNN \cite{kim2019edge} & No  &  66.85  \\
  \textbf{Ours} & No & \textbf{68.35} \\
\hline
 MAML \cite{finn2017model}  & BN  &  63.11  \\
  Reptile + BN \cite{nichol2018first}  & BN  &  65.99  \\
  Relation Net \cite{sung2018learning} & BN  &  67.07 \\
  \hline
  MAML + Transduction \cite{finn2017model}  & Yes  &  66.19  \\
  TNP \cite{liu2019learning} & Yes & 69.43 \\
  TPN (Higher K) \cite{liu2019learning} & Yes & 69.86\\
  EGNN+Transduction \cite{kim2019edge} & Yes & 76.37\\
  \textbf{Ours+Transduction} & Yes & \textbf{77.78}\\
  \hline
\end{tabular}
\caption{\small Results of 5-way 5-shot learning on {\it mini}ImageNet, averaged
	over 600 test episodes.
``Ours'' means EGNN plus our TMDlayer.``BN'' means that the query batch
statistics are used instead of global batch normalization parameters.}
\label{table:few_shot}
\end{table}

\begin{table}[!h]\small

\centering
\begin{tabular}{l  | c c c c}
\hline
 & \multicolumn{4}{c}{Labeled Ratio (5-way 5-shot)}\\
 Training method & 20\% & 40\% & 60\% & 1000\% \\
\hline
GNN-semi \cite{garcia2017few} & 52.45& 58.76 &-& 66.41  \\
 EGNN-semi \cite{kim2019edge} & 61.88 & 62.52 & 63.53 & 66.85  \\
 \textbf{Ours} & \textbf{63.14} & \textbf{64.32} & \textbf{64.83} & 68.35  \\
\hline
 EGNN-semi(T) \cite{kim2019edge} & 63.62 & 64.32 & 66.37 & 76.37  \\
 \textbf{Ours(T)} & \textbf{64.84} & \textbf{66.43} & \textbf{68.62} & 77.78  \\
 \hline
\end{tabular}
\caption{\small Accuracy  of semi-supervised few-shot classification. ``Ours'' means EGNN plus our TMDlayer.}
\label{table:semi_few_shot}
\end{table}

%% file: conclusion.tex
{\subsection{Runtime overhead/Relation with augmentation}
\textbf{Runtime overhead.} Our construction does involve some {\em training time} overhead because of computing the kernel matrix, and varies
depending on the use case. For reference, the overhead is $10\%$ in \S\ref{sec:point_cloud}, $11\%$ in \S\ref{sec:object_segmentation} and $1\%$ in \S\ref{sec:few_shot}.

\textbf{Relationship with data augmentation.} Data augmentation and TMDLayer are complementary, not mutually exclusive. In all our experiments,
the baselines use data augmentations (e.g., random rotation or cropping). Our TMDLayer offers benefits, above and beyond augmentation. 
}

\section{Discussion and Conclusions} We proposed an SDE based framework that allows a unified view of several different learning tasks in vision.
Our framework is beneficial where data generation (or the data itself) can be described using stochastic processes, or more specifically diffusion operators.
This is particularly useful in settings where obtaining a deterministic model of the image manifold or
learning density functions are impossible or challenging due to high sample complexity requirements.
Our TMDlayer does {\em not} require explicit generation of diffused samples, especially during training, making it computationally efficient. 
The ``process'' of which the provided data sample is a snapshot and whose characterization is enabled by our TMDlayer,
also appears to have implications for robust learning. Indeed, if the 
parameters that define the process are explicitly optimized, 
we should be able to establish an analogy between the resultant model as a stochastic/simpler version of recent results 
for certified margin radius maximization \cite{zhen2021simpler} which often require access to Monte Carlo sampling
oracles \cite{cohen2019certified}. We believe that periodicity in
SDEs for data augmentation is an important missing ingredient -- for instance --  this may help model seasonal patterns
in disease progression studies for predictions, automatically. For this purpose, tools from Floquet theory may
allow us to consider transformed versions of the process, potentially with simplified generators.
{Our code is available at \href{https://github.com/zihangm/neural-tmd-layer}{https://github.com/zihangm/neural-tmd-layer} }.

\section*{Acknowledgments} This work was supported by NIH grants RF1 AG059312
and RF1 AG062336. SNR was supported by UIC start-up
funds. We thank Baba Vemuri for providing many important suggestions on formulating the Chan-Vese model within deep networks. 

